\documentclass{article}
\usepackage{ijcai24}

\usepackage{times}
\usepackage{soul}
\usepackage{url}
\usepackage[hidelinks]{hyperref}
\usepackage[utf8]{inputenc}
\usepackage[small]{caption}
\usepackage{graphicx}
\usepackage{amsmath}
\usepackage{amsthm}
\usepackage{booktabs}
\usepackage{algorithm}
\usepackage{algorithmic}
\usepackage[switch]{lineno}

\usepackage{amssymb}
\usepackage{wrapfig}

\DeclareMathOperator*{\argmin}{arg\,min}

\newtheorem{theorem}{Theorem}

\theoremstyle{definition}
\newtheorem{definition}[theorem]{Definition}

\title{Towards Exact Computation of Inductive Bias}

\author{
Akhilan Boopathy\and
William Yue\and
Jaedong Hwang\and
Abhiram Iyer\and
Ila Fiete\\
\affiliations
Massachusetts Institute of Technology
\emails
akhilan@mit.edu
}

\begin{document}

\maketitle

\begin{abstract}
Much research in machine learning involves finding appropriate inductive biases (e.g. convolutional neural networks, momentum-based optimizers, transformers) to promote generalization on tasks. However, quantification of the amount of inductive bias associated with these architectures and hyperparameters has been limited. We propose a novel method for efficiently computing the inductive bias required for generalization on a task with a fixed training data budget; formally, this corresponds to the amount of information required to specify well-generalizing models within a specific hypothesis space of models. Our approach involves modeling the loss distribution of random hypotheses drawn from a hypothesis space to estimate the required inductive bias for a task relative to these hypotheses. Unlike prior work, our method provides a direct estimate of inductive bias without using bounds and is applicable to diverse hypothesis spaces. Moreover, we derive approximation error bounds for our estimation approach in terms of the number of sampled hypotheses. Consistent with prior results, our empirical results demonstrate that higher dimensional tasks require greater inductive bias. We show that relative to other expressive model classes, neural networks as a model class encode large amounts of inductive bias. Furthermore, our measure quantifies the relative difference in inductive bias between different neural network architectures. Our proposed inductive bias metric provides an information-theoretic interpretation of the benefits of specific model architectures for certain tasks and provides a quantitative guide to developing tasks requiring greater inductive bias, thereby encouraging the development of more powerful inductive biases.
\end{abstract}

\section{Introduction}
Generalization is a fundamental challenge in machine learning, as models must be able to perform well on unseen data after being trained on a limited set of examples. To achieve this, researchers have extensively studied the role of inductive biases, which are prior assumptions or restrictions embedded within learning algorithms, in promoting generalization. These biases can take various forms, such as architectural choices (e.g., convolutional neural networks, momentum-based optimizers, transformers) or hyperparameter settings, and they shape the space of hypotheses that the model can consider.

Despite the importance of inductive biases, quantifying the amount of inductive bias associated with different architectural and hyperparameter choices has remained challenging. Inductive bias can be formulated as the amount of information required to specify well-generalizing models within a \textit{hypothesis space} of models~\cite{chollet2019arc,boopathy2023model}. Like a model class, a hypothesis space is a set of models; however, while a model class corresponds to a specific set of inductive biases (e.g. a particular architecture), a hypothesis space is a set in which \textit{all} relevant model classes are contained. We make a distinction between hypothesis spaces and model classes to illustrate that hypothesis spaces are typically much broader than model classes and set the context under which inductive biases can be evaluated; for instance, a model class may be a particular convolutional neural network architecture, while the hypothesis space may consist of \textit{all} functions expressible by any finite-sized neural network.

Previous attempts at measuring inductive bias have often provided only upper bounds or have been limited to specific model classes. This limitation hinders a comprehensive understanding of how different biases contribute to generalization and impedes the systematic development of more effective biases.

In this paper, we propose a novel and efficient method for computing the inductive bias required for generalization on a task under fixed training data budget. Unlike prior work, our approach provides a direct estimate of inductive bias without relying on bounds. Moreover, it is applicable to diverse hypothesis spaces, allowing the computation of inductive bias \textit{within} the context of particular model classes such as neural networks. We believe more precise and flexible computation of inductive bias is practically valuable:

First, by quantifying the amount of inductive bias associated with different architectural choices, researchers can gain profound insights into how specific design decisions affect the model's ability to generalize. This understanding helps identify which architectural features contribute most significantly to improved performance and informs the development of more tailored and task-specific models. Armed with a quantitative measure of inductive bias, practitioners can make more informed decisions about which architectural choices to prioritize when building and optimizing machine learning models. This, in turn, can lead to more efficient model development processes and improved real-world applications.

Second, our inductive bias measure serves as a practical guide for designing tasks that demand higher levels of inductive bias. By precisely estimating the amount of inductive bias needed for a given task, researchers can intentionally craft benchmarks that challenge the boundaries of generalizability of current models. This approach encourages the development of more powerful model architectures and learning algorithms, fostering innovation in the field.

We summarize our contributions as follows:\footnote{\url{https://github.com/FieteLab/Exact-Inductive-Bias}}
\begin{itemize}
    \item We propose a definition of inductive bias with an explicit dependence on the hypothesis space within which models are defined.
    \item We develop an efficient sampling-based algorithm to compute the inductive bias required to generalize on a task. Unlike prior work, the method can be applied to parametric and non-parametric hypothesis spaces.
    \item We derive an upper bound on the approximation error of inductive bias estimate; the approximation error scales inversely with the number of sampled hypotheses.
    \item We empirically apply our inductive bias metric to a range of domains including supervised image classification, reinforcement learning (RL) and few-shot meta-learning. Consistent with prior work, we find that tasks with higher dimensional inputs require more inductive bias.
    \item We empirically find that neural networks encode massive amounts of inductive bias relative to other expressive model classes. Furthermore, we quantify the difference in inductive bias provided by different neural network architectures within a neural network hypothesis space.
\end{itemize}

\section{Related Work}

\paragraph{Generalization vs. Sample Complexity}

Traditionally, the generalizability of machine learning models has been analyzed in terms of sample complexity, which is the amount of training data required to generalize on a task~\cite{cortes1994learning,murata1992learning,amari1993universal,hestness2017deep}. Measures such as Rademacher complexity~\cite{koltchinskii2000rademacher} and VC dimension~\cite{blumer1989learnability} quantify the capacity of a model class and provide upper bounds on sample complexity, with less expressive model classes requiring fewer samples. More recently, data-dependent generalization bounds have been proposed, yielding tighter bounds based on dataset properties~\cite{negrea2019information,raginsky2016information,kawaguchi2022robustness,lei2015multi,jiang2021methods}. Additionally, scaling laws for neural networks have modeled learning as kernel regression, revealing that sample complexity scales exponentially with the intrinsic dimensionality of data~\cite{bahri2021explaining,hutter2021learning,sharma2022scaling}. In our work, instead of focusing on the importance of training data in generalization, we focus on the role of inductive biases.

\paragraph{Generalization vs. Inductive Bias Complexity}

The importance of inductive biases in promoting generalization has been widely recognized, starting with the No Free Lunch theorem~\cite{wolpert1996lack} which states that no learning algorithm can perform well on \textit{all} possible tasks: learning algorithms require inductive biases tailored to specific sets of tasks. Subsequent studies have further emphasized the role of inductive biases in learning~\cite{hernandez2016evaluation,haussler1988quantifiying,du2018samples,li2021convolutional}, showing that specific abilities, biases, and model architectures provide prior knowledge that facilitates generalization. Despite the central role of inductive biases, work on quantifying them has been limited. \cite{chollet2019arc} proposes measuring the generalization difficulty of a task as the amount of inductive bias required for a learning system to perform the task in addition to any training data provided. \cite{boopathy2023model} provides an upper bound on the \textit{inductive bias complexity} of a task (i.e. how much inductive bias is required to generalize on a task) based on task properties. In particular, it finds that higher-dimensional tasks (i.e. tasks with inputs of higher intrinsic dimensionality) require exponentially greater inductive bias, mirroring results for sample complexity. In this work, we aim to more precisely and directly estimate the required inductive bias of a task without the use of bounds. Moreover, unlike prior work, our approach can compute inductive bias complexity within general hypothesis spaces: it allows for \textit{context-specific} computation of inductive bias. For instance, we may compute the inductive bias required to generalize on ImageNet~\cite{deng2009imagenet} classification under 1) a hypothesis space of general neural networks vs. 2) a hypothesis space consisting only of convolutionally-structured neural networks, under which less inductive bias is required.

\begin{figure}[t!]
    \centering
    \includegraphics[width=0.8\linewidth]{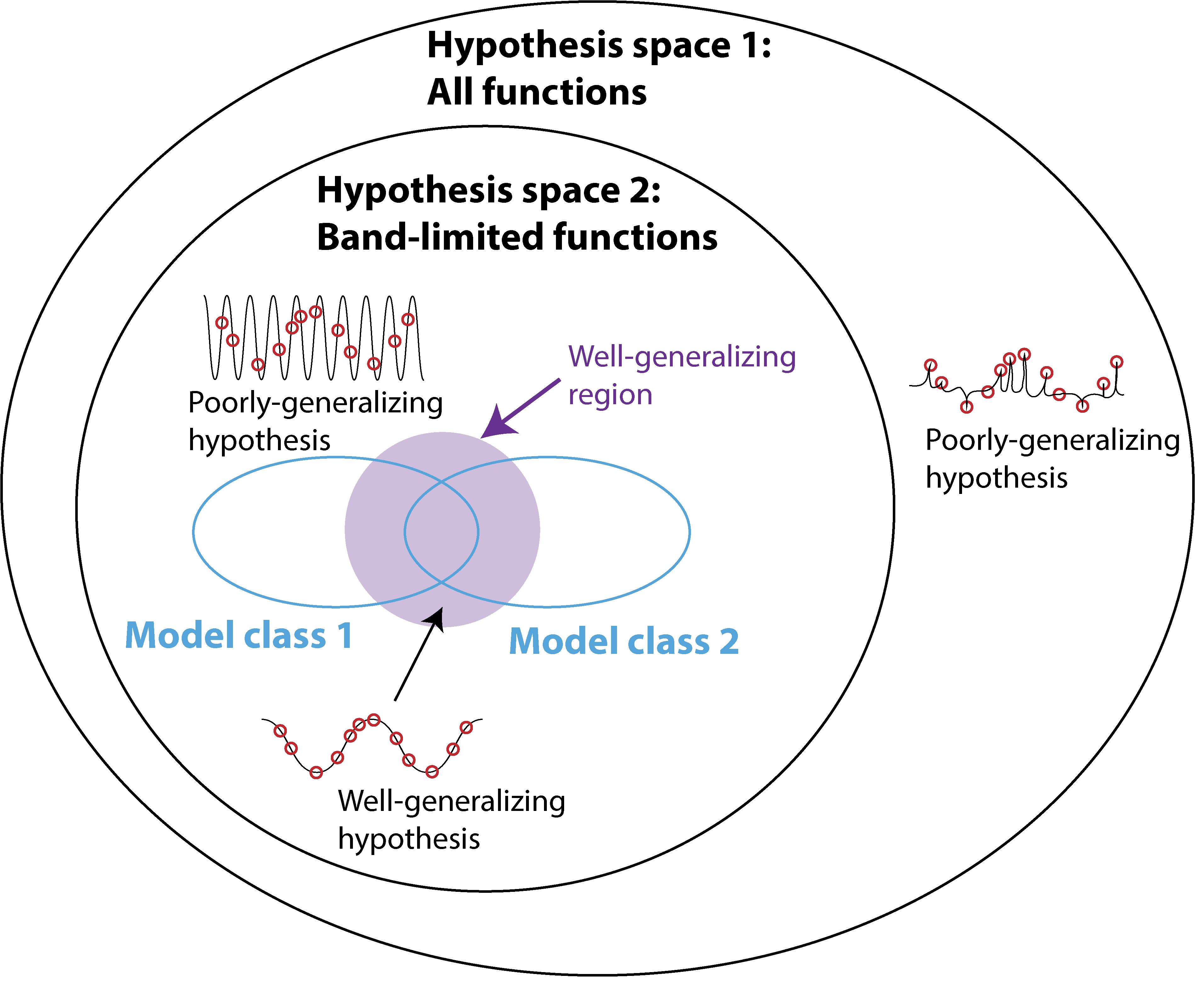}
    \caption{An illustration of example hypothesis spaces, model classes, and specific models for a particular learning problem. Red circles indicate training points and black curves indicate hypotheses. A hypothesis space sets the broad set of models we wish to consider. In this illustration, we consider the hypothesis space of all functions and a smaller hypothesis space of band-limited functions (i.e. functions with limited maximum frequency). A model class is a set of models associated with a particular set of inductive biases. We measure the required amount of inductive bias to solve a task based on the size of the well-generalizing region within the context of a particular hypothesis space.  }
    \label{fig:hypothesis_space_diagram}
\end{figure}

\section{Quantifying Inductive Bias}
In this section, we 
present our method for quantifying the inductive bias of a model class. Inductive bias, in simple terms, represents the inherent assumptions or characteristics of a model class that influence its ability to generalize to new, unseen data. We first provide a formal quantitative definition of the amount of inductive bias of a model class. We then propose our method of approximating this amount of inductive bias and prove a bound on its error.

\begin{figure*}[t!]
    \centering
\includegraphics[width=0.7\textwidth]{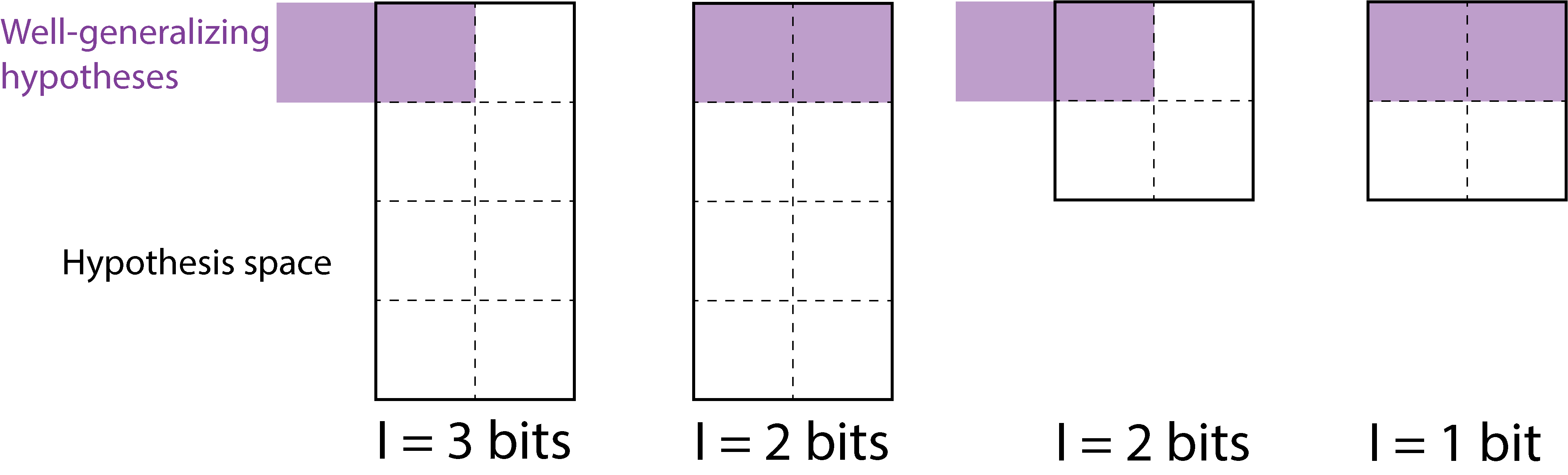}
    \caption{Illustration of how the required inductive bias for a task can be computed from the hypothesis space and the region of well-generalizing hypotheses. Black boxes indicate hypothesis spaces; $p_h$ is a uniform distribution over each box. Purple indicates regions of well-generalizing hypotheses. Inductive bias is the negative log of the fraction of hypothesis space that generalizes well: $I = -\log \frac{Hypothesis\,space \cap Well-generalizing\,hypotheses}{Hypothesis\,space}$. It depends on both the size of the hypothesis space as well as how much the hypothesis space overlaps with well-generalizing hypotheses. Different hypothesis spaces may yield different inductive bias estimates \textit{even on the same task} (i.e. the same set of well-generalizing hypotheses).}
    \label{fig:hypothesis_space}
\end{figure*}

\subsection{Definition of Inductive Bias}
Intuitively, inductive biases are guiding principles that help models make sense of data. For instance, when we look at an image of a cat, we rely on our inductive bias to recognize it as a cat, based on features like whiskers, fur, and ears. In machine learning, model classes (e.g. convolutional neural networks) have inductive biases (e.g. invariance to translation of inputs) which are essential for sample-efficient learning.

\cite{boopathy2023model} proposes quantifying the amount of inductive bias required to generalize on a task based on the probability that a model that fits a training set also generalizes to a test set. Importantly, this definition assumes that there exists a \textit{hypothesis space} of models. Formally, the hypothesis space contains all possible models to solve a specific problem (i.e. the universe of potential solutions). A model is a specific hypothesis from this space. A \textit{model class} is a subset of this space, chosen by a model designer, typically including well-generalizing models. Inductive bias, then, quantifies the amount of information needed to specify a well-generalizing model class within the hypothesis space. In simpler terms, it measures how much "guidance" a model needs from the model designer to perform well on a task. See Figure~\ref{fig:hypothesis_space_diagram} for an illustration of these concepts.

Note that the size of the hypothesis space can strongly affect the magnitude of the inductive bias, but in \cite{boopathy2023model} the dependence on the hypothesis is implicit. Here we formally define inductive bias in a similar manner to~\cite{boopathy2023model} but provide a way to explicitly include the distribution of models in the hypothesis space (e.g. neural networks versus Gaussian RBF models; or more finely, different neural network architectures). Our approach applies across a variety of domains ranging from supervised classification to RL as we will empirically show. Our formal definition follows:

\begin{definition}
    Let $\mathcal{H}$ be a set of hypotheses, and let hypothesis distribution $p_h$ define a probability distribution over these hypotheses. Suppose there exists a loss function $L$ that maps a hypothesis $h \in \mathcal{H}$ and a task input $x \in \mathcal{X}$ to a scalar: $L: \mathcal{H} \times \mathcal{X} \to \mathbb{R}$. Finally, suppose there exists a test distribution $p_x$ over the task inputs. With respect to distribution $p_h$, the amount of inductive bias required to achieve test set error rate $\varepsilon$ on a task is:
    \begin{equation} \label{eqn:definition}
        I(\varepsilon, p_h, p_x, L) = -\log \int \mathbf{1}(\mathbb{E}_{x \sim p_x}[L(h, x)] \leq \varepsilon) p_h(h) dh
    \end{equation}
    where $\mathbf{1}$ denotes the indicator function.
\end{definition}
Note that this is simply negative log of the probability that a hypothesis sampled from $p_h$ achieves an error rate $\leq \varepsilon$ on a test set. $p_h$ may be any distribution over the hypothesis space; \cite{boopathy2023model} sets $p_h$ as a uniform distribution of models achieving a training set error $\leq \epsilon$. In practice, we may be interested in the case when $p_h$ is a distribution of models produced by an optimization process on a training set. This allows us to quantify the \textit{additional} inductive bias required to generalize on top of any information provided by the training data. Critically, as Figure~\ref{fig:hypothesis_space} illustrates, the specific choice of $p_h$ has a significant impact on the inductive bias. Intuitively, if the hypothesis distribution is more aligned with a task, fewer inductive biases are required to generalize.

Estimating this inductive bias by directly sampling hypotheses from a hypotheses space is computationally infeasible for large hypothesis spaces since the vast majority of hypotheses may not generalize well. Thus, we propose a two-phase approach to compute the inductive bias: first, we sample from the hypothesis space and compute an empirical distribution of test set error values $\mathbb{E}_{x \sim p_x}[L(h, x)]$. Few (or none) of these hypotheses may generalize at the desired error rate. Thus, we use the samples to model the test error distribution to estimate the probability of achieving test error $\leq \varepsilon$.

We also note that inductive bias in Equation~\ref{eqn:definition} is a function of the desired error rate $\varepsilon$; it is not a function of a specific model or model class, although it is a function of the hypothesis distribution $p_h$. However, we may use this definition to compute the inductive bias \textit{provided} by a specific model under a specific hypothesis space by computing the amount of inductive bias required to generalize at the level of the model (i.e. by plugging in the model's test set error rate $\varepsilon$ into Equation~\ref{eqn:definition}). This allows us to understand how the inductive bias of a model is affected by the properties of the broader hypothesis space, and how it contributes to the model's generalization.

\subsection{Efficiently Sampling from the Hypothesis Space}
Here, we aim to efficiently sample hypotheses from $p_h$, where we assume $p_h$ includes only hypotheses fitted to a training set. We use two approaches: directly optimizing the parameters of a hypothesis (i.e. training a model on the training data), or a kernel-based sampling approach.

\paragraph{Direct Optimization by Gradient Descent}
For hypothesis spaces with a known, finite-dimensional parameterization, it may be reasonable to set $p_h$ as a distribution of hypotheses produced by performing gradient descent on loss function $L$ evaluated on a training set of data $x$. For instance, $p_h$ may correspond to a distribution of neural networks after training from random initialization by gradient descent on a training set. Given $P$ parameters per hypothesis, performing each step of gradient descent takes $O(P)$ time, yielding $O(PT)$ time for $T$ optimization steps. Thus, producing $S$ samples requires $O(SPT)$ time.

\paragraph{Kernel-based Sampling}
If the hypothesis space is very high-dimensional, direct optimization may be computationally challenge, and for infinite-dimensional models, representing the parameters themselves may be infeasible. Instead, we formulate the problem of sampling from a hypothesis space as sampling from a Gaussian process, for which efficient algorithms have been extensively studied. We use an algorithm resembling the approach of~\cite{lin2023sampling}. The key principle behind our algorithm is to reparameterize the distribution of hypothesis output values on a test set in terms of a unit Gaussian. This allows us to easily and efficiently draw samples from this distribution in linear time (in terms of training set size).

We assume hypotheses $h$ are linearly parameterized with parameters $\theta \in \mathbb{R}^P$ as:
\begin{equation}
    h(x) = \phi(x) \theta
\end{equation}
where $\phi(x) \in \mathbb{R}^{k \times P}$ is a dimensional feature matrix and $k$ is the dimensionality of $h(x)$. This type of assumption is standard for kernel methods including neural networks in the Gaussian process or Neural Tangent Kernel
limits. Here, we set $p_h$ to include only the set of hypotheses that \textit{interpolate} the training data. Given a set of $N$ training points $X$, their corresponding features $\phi(X) \in \mathbb{R}^{N k \times P}$ and target model outputs $Y \in \mathbb{R}^{N k}$, where $k$ represents output dimensionality, observe that if a hypothesis interpolates the training data, its parameters must satisfy:
\begin{equation}
    Y = \phi(X) \theta
\end{equation}
We may decompose $\theta$ into two terms:
\begin{equation}
    \theta = \phi(X)^{\dagger} Y + \beta
\end{equation}
where $\beta \in \mathbb{R}^P$ satisfies $\phi(X) \beta = 0$. The first term ensures the hypothesis fits the training data while the second term allows for variation between hypotheses. Finally, we set $\beta$ as a Gaussian with mean $0$ and covariance $I - \phi(X)^\dagger \phi(X)$, where $\dagger$ represents pseudoinverse. This corresponds to setting the distribution of parameters $\theta$ as:
\begin{equation}
    p_\theta(\theta) = \mathcal{N}(\phi(X)^{\dagger} Y, I - \phi(X)^\dagger \phi(X))
\end{equation}
This corresponds to a Gaussian process conditioned on the training points.

We aim to sample the value of $h$ on a test set $\bar X$ consisting of $n$ points. These values $h(\bar X)$ may be computed as:
\begin{equation}
    h(\bar X) = K(\bar X, X) \alpha^* + \sqrt{K(\bar X, \bar X) - K(\bar X, X) A^*} z
\end{equation}
where $\alpha^*$ and $A^*$ are found as:
\begin{equation}
    \alpha^* = \argmin_\alpha ||Y - K(X, X) \alpha||_2^2
\end{equation}
\begin{equation}
    A^* = \argmin_A ||K(X, \bar X) - K(X, X) A||_F^2
\end{equation}
and $z$ is drawn from a unit Gaussian $\mathcal{N}(0, I)$. We find approximate solutions to these optimization problems by stochastic gradient descent. Pseudocode is provided in Algorithm~\ref{alg:kernel_sampling}, with additional details included in the Appendix. Producing $S$ samples requires a total of $O(nNk^2T + n^3k^3 + n^2Nk^3 + n^2k^2S)$ time where $T$ is the number of optimization steps. In practice, the $nNk^2T$ term dominates due to the large size of $T$: this implies that we may increase the number of drawn samples $S$ with $\textit{no asymptotic impact}$ on the runtime.

\begin{algorithm}
\caption{Kernel-based Sampling}\label{alg:kernel_sampling}
\begin{algorithmic}[1]
\REQUIRE{$X, Y, \bar X, n, k, T, S, \eta$}
  \STATE{Initialize $\alpha$ and $A$ with zeros}
  \FOR{$t = 1$ to $T$}
    \STATE Randomly sample a mini-batch of training examples $(x, y) \in (X, Y)$
    \STATE $g_\alpha = 2 K(X, x) (K(x, X) \alpha - y)$
    \STATE $g_A = 2 K(X, x) (K(x, \bar X) A - K(x, \bar X))$
    \STATE $\alpha \gets \alpha - \eta g_\alpha$
    \STATE $A \gets A - \eta g_A$
  \ENDFOR
  \STATE $m = K(\bar X, X)\alpha$
  \STATE $\sqrt{C} = \sqrt{K(\bar X, \bar X) - K(\bar X, X) A}$
  \STATE \textbf{Initialize} an empty list $\mathtt{samples}$
  \FOR{$s = 1$ to $S$}
    \STATE Sample $z \sim \mathcal{N}(0, I)$
    \STATE $h(\bar X) = m + \sqrt{C}z$
    \STATE Append $h(\bar x_s)$ to $\mathtt{samples}$
  \ENDFOR
\STATE  Return $\mathtt{samples}$
\end{algorithmic}
\end{algorithm}

\subsection{Modeling the Test Error Distribution} \label{chi-squared}
Once we generate samples from the hypothesis space, we next aim to model the distribution of test set losses of sampled functions from the hypothesis space; this allows us to compute inductive bias.

To understand the shape of the test loss distribution, we return to the assumptions in the kernel-based sampling of the hypothesis space. We assume that the regression targets $Y, \bar Y$ on the training and test sets respectively are constructed as $Y = \phi(X) \theta^*, \bar Y = \phi(\bar X) \theta^*$ for unknown parameters $\theta^*$.

The squared error between the prediction $h(\bar X)$ and true value $\bar Y$ may be written as:
\begin{multline}
||h(\bar X) - \bar Y||_2^2 \\= ||\phi(\bar X) \phi(X)^{\dagger} \phi(X) \theta^* + \phi(\bar X) \beta - \phi(\bar X) \theta^*||_2^2
\end{multline}
We may write $\beta$ as $\beta = (I - \phi(X)^\dagger \phi(X)) \xi$ where $\xi \in \mathbb{R}^{P}$ is distributed as a unit Gaussian. Then, the squared prediction error may be written as:
\begin{equation}
||h(\bar X) - \bar Y||_2^2 = ||\phi(\bar X) (I - \phi(X)^\dagger \phi(X)) (\xi - \theta^*)||_2^2
\end{equation}
Observe that this is a quadratic form of a Gaussian random variable $\xi - \theta^*$; thus, $||h(\bar X) - \bar Y||_2^2$ follows a generalized Chi-squared distribution.

\begin{figure}
    \centering
    \includegraphics[width=0.9\linewidth]{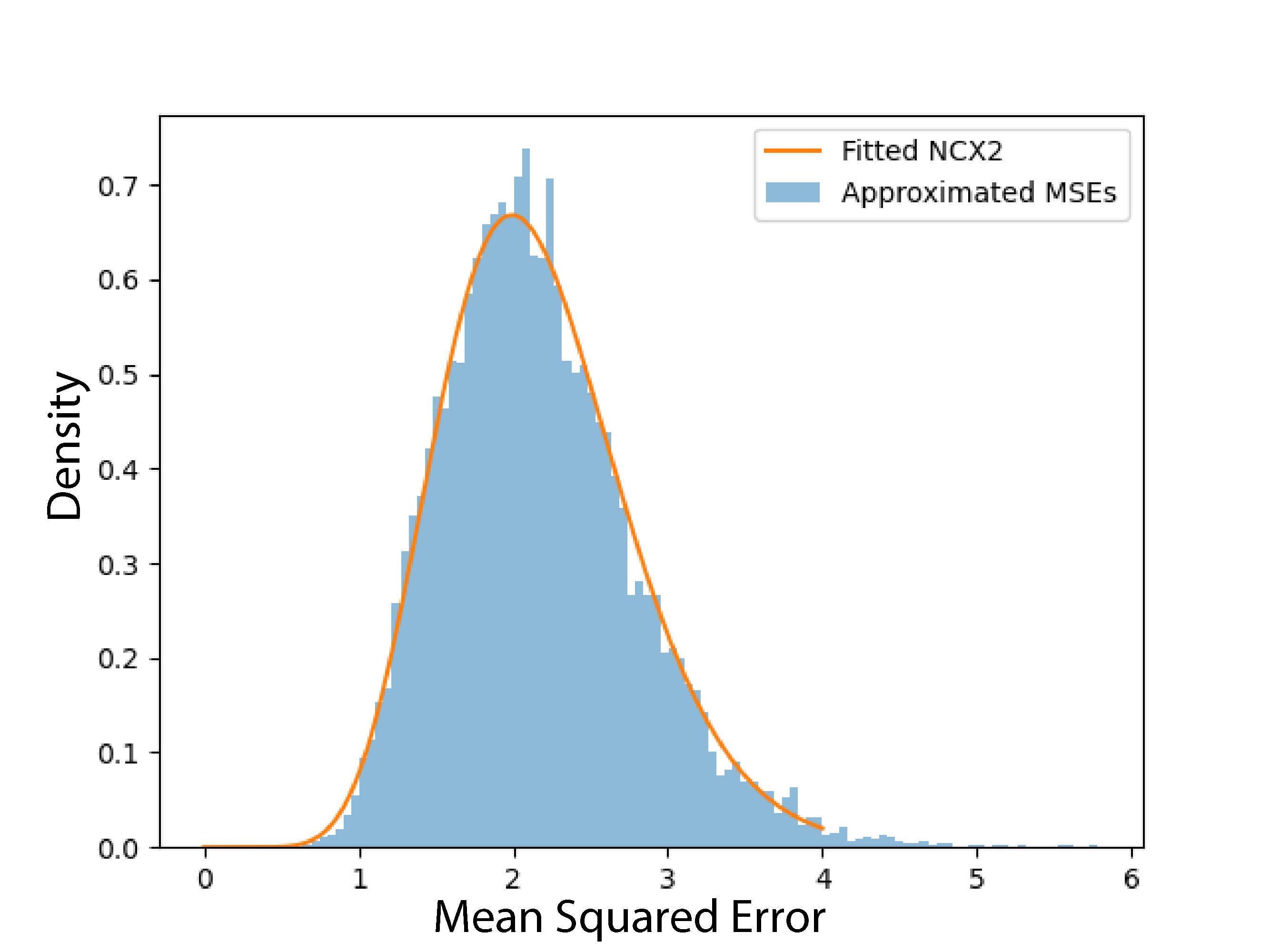}
    \caption{Fitting a scaled non-central Chi-squared distribution to an empirical distribution of mean squared errors of models drawn from a kernel-based Gaussian RBF hypothesis space on a restricted version of MNIST. Observe that the distribution closely models the empirical distribution.}
    \label{fig:dist_fit}
\end{figure}

In practice, to minimize the number of fit parameters when modeling the empirical error distribution, we fit the test loss using a scaled non-central Chi-squared (which is a special case of a generalized Chi-squared distribution); this has three fit parameters. These parameters are fit with maximum likelihood estimation. Figure~\ref{fig:dist_fit} illustrates that this distribution is able to closely fit test errors of random hypotheses on a real dataset. 
Since we model the test error distribution as a Chi-squared distribution, we need to approximate the negative log of the cumulative distribution function (CDF) given its parameters. Given a Chi-squared distribution with $k$ degrees of freedom and non-centrality parameter $\lambda$, we use the following approximation by Sankaran~\cite{sankaran1959non} for the CDF:
\begin{multline}
    P(z;k,\lambda) \\ \approx\Phi\left[\frac{\left(\frac{z}{k+\lambda}\right)^h-(1+hp(h-1-0.5(2-h)mp))}{h\sqrt{2p}(1+0.5mp)}\right],
\end{multline}
where $\Phi$ is the CDF of a standard normal random variable and
\begin{multline}
h=1-\frac{2}{3}\frac{(k+\lambda)(k+3\lambda)}{(k+2\lambda)^2},\qquad p=\frac{k+2\lambda}{(k+\lambda)^2},\\ m=(h-1)(1-3h).
\end{multline}
This approximation is more accurate when $k$ is large, which corresponds to higher-dimensional parameter spaces; for lower dimensional spaces, it may be more appropriate to directly compute the CDF of the Chi-squared distribution. Finally, we use a Chernoff bound-based approximation $\Phi(-z)\approx e^{\frac{-z^2}{2}}$ to finish the calculation. Specifically, since the inductive bias is given as the negative log probability of generalizing up to error rate $\varepsilon$ (see Equation~\ref{eqn:definition}), we simply compute the negative log of the approximated CDF after plugging in the desired $\varepsilon$ for $z$. The appendix provides further details on how the test error distribution is modeled.

\subsection{Bounding the Approximation Error}
Next, we derive a bound on the approximation error of our estimate of required inductive bias. At a high level, the bound proceeds as follows: we first bound how closely our samples from the hypothesis distribution match the true distribution $p_h$, and then bound the error in our modeling of the test error distribution to arrive at a final bound on the amount of inductive bias.
\begin{theorem}
    Suppose we are provided a hypothesis distribution $p_h$, input distribution $p_x$, loss function $L$ and desired error rate $\varepsilon$. Suppose we estimate $I(\varepsilon, p_h, p_x, L)$ by first sampling $n$ hypotheses ($h^1, h^2, ... h^n$) iid from $q_h$ which is close to $p_h$ in the sense that 
    \begin{equation}
        |\log p_h(h) - \log q_h(h)| \leq \xi_h
    \end{equation} 
    for all $h$. We then compute the test losses of each hypothesis $\mathbb{E}_{x \sim p_x}[L(h^1, x)], \mathbb{E}_{x \sim p_x}[L(h^2, x)], ... \mathbb{E}_{x \sim p_x}[L(h^n, x)]$. Next, we model the distribution of test losses with a distribution $f(l;\alpha)$ where $\alpha$ represents a finite number of parameters. We assume $f$ has bounded support over $l$. We assume that knowing a finite number of moments of $f$ uniquely determines $\alpha$ in the sense that there exists $\tilde f$ such that $\tilde f(l;\mu) = f(l;\alpha)$ where $\mu$ represent $r$ moments of the distribution:
    \begin{equation}
        \mu = \int M(l)  f(l;\alpha) dl
    \end{equation}
    for some function $M(l)$. We assume $\log \tilde f$ is Lipschitz continuous with respect to $\mu$.
    
    Denote the distribution of $\mathbb{E}_{x \sim p_x}[L(h, x)]$ when $h$ is drawn from $q_h$ as $q_l$. We assume $q_l$ can be closely modeled by $f(l;\alpha)$ in the following sense:
    \begin{equation}
        \max_l |\log q_l(l) - \log \tilde f(l;\bar \mu)| \leq \xi_l
    \end{equation}
    where $\bar \mu = \int M(l) q_l(l) dl$ are the moments of $q_l$. Given the empirical test loss distribution, we use the method of moments to estimate the parameters of $f$, yielding $\alpha^*$. Finally, suppose that the estimate of $I(\varepsilon, p_h, p_x, L)$ is computed as:
    \begin{equation}
        \tilde I = -\log \int_{-\infty}^{\varepsilon} f(l;\alpha^*) dl
    \end{equation}
    Then with probability $1-\sigma$, the approximation error of $\tilde I$ can be bounded as:
    \begin{equation}
        | \tilde I - I(\varepsilon, p_h, p_x, L)| \leq \xi_h + \xi_l + \frac{\kappa}{n} \sqrt{r \log \frac{2r}{\sigma}}
    \end{equation}
    for a constant $\kappa$.
\end{theorem}
See the appendix for a proof. Observe that the approximation error is bounded by three terms: the first corresponds to how accurately the hypothesis distribution can be sampled, the second corresponds to the modeling error of the test error distribution, and the third corresponds to the error from drawing a finite number of samples. Practically, the first term $\xi_h$ can be set to $0$ if we are able to sample from $p_h$. Similarly, $\xi_l=0$ is $0$ if the test error distribution follows a scaled non-central Chi-squared distribution, which can be motivated theoretically and empirically as explained in Section~\ref{chi-squared}. Thus, the remaining error is the finite sample approximation error which converges with rate $O(\frac{1}{n})$. We also add that when $p_h, p_l, q_h, q_l$ are known explicitly, $\xi_h$ and $\xi_l$ can typically be bounded explicitly as well (e.g. when the hypothesis or loss distributions are discrete with finite support). 

Note that instead of the method of moments, we use maximum likelihood estimation to estimate $\alpha$ since it is practically effective. Maximum likelihood estimation can also be shown to yield the same convergence rate asymptotically with $n$, although deriving a finite sample bound is more challenging.

\section{Experimental Results}
We first evaluate the amount of inductive bias required to generalize on various tasks under different choices of hypothesis space and compare our estimates to prior work. We then use our approach to assess the amount of inductive bias \textit{contained} in different models.

\subsection{Inductive Bias Required to Generalize on Different Tasks}
\begin{table}[t!]
    \centering
    \resizebox{0.47\textwidth}{!}{
    \begin{tabular}{cccc}
        \toprule
        Task & Upper Bound~\cite{boopathy2023model} & Our Results (kernel) & Our Results (NN)\\
        \midrule
        Inverted Pendulum & $4.41\times 10^9{*}$ & $29$ & $1.0$\\
        MNIST & $1.48\times 10^{16}$  & $2568$ & $37.6$\\
        CIFAR-10 & $3.43\times 10^{32}$  & $2670$ & $327.3$\\
        Omniglot & $1.79\times 10^{145}$  & $2857$ & $1925.6$\\
        \bottomrule
    \end{tabular}
    }
    \caption{Our inductive bias estimates (bits) for both Gaussian RBF kernel and neural network (NN) hypothesis spaces on image classification datasets (MNIST, CIFAR-10), Omniglot, and Inverted Pendulum tasks. We compare with bounds from \protect\cite{boopathy2023model} *Our version of the Inverted Pendulum task differs somewhat from \protect\cite{boopathy2023model}.}
    \label{tab:bitsresults}
\end{table}
Our evaluation includes benchmark tasks across various domains: MNIST~\cite{lecun1998mnist}, CIFAR-10~\cite{krizhevsky2009cifar}, 5-way 1-shot Omniglot~\cite{lake2015HumanlevelCL} and inverted pendulum control~\cite{florian2007correct}. We treat classification tasks as regression problems using mean squared error loss with one-hot-encoded labels. Two hypothesis spaces are examined: a Gaussian RBF kernel-based space and a high-capacity ReLU-activated neural network.

The kernel-based hypothesis space uses a Gaussian RBF kernel constructed as $K(x_1, x_2) = e^{-\frac{1}{2} ||x_1 - x_2||_2^2 I}$. Hypotheses from this space are constrained to fit the training data; thus, our inductive bias measure quantifies additional information required to generalize on top of the training data. We use Algorithm~\ref{alg:kernel_sampling} to sample hypotheses from this space and evaluate their mean squared error on the test set of each task. Note that this hypothesis space is infinite-dimensional; directly optimizing in the space is not feasible. Due to computational constraints, gradient descent in Algorithm~\ref{alg:kernel_sampling} is not run until full convergence; thus, sampled hypotheses may not interpolate the training data. Nevertheless, we find that the corresponding distribution of losses stabilizes after a small number of epochs (see the Appendix). 
\begin{table*}[t!]
    \centering
    \resizebox{0.75\textwidth}{!}{
    \begin{tabular}{ccccccccc}
    \toprule
    & Linear & FC & Deep CNN & Alexnet & LeNet-5 & DSN & MCDNN & DenseNet \\
    \midrule
    MNIST & 0.0 & 0.0 & - & 0.0 & 0.0 & 0.3 & 7.7 & - \\
    CIFAR-10 & 0.0 & 0.0 & - & 0.0 & - & 0.0 & 0.0 & 7.2 \\
    SVHN & - & 0.0 & 345.7 & 67.5 & - & 405.6 & - & 507.1 \\
    \toprule
    & Matching Nets & MAML  & Prototypical Nets & iMAML & MT-net & & & \\
    \midrule
    Omniglot & 1400  & 1700 & 1800 & 2500 & 2500  & & & \\
    \bottomrule
    \end{tabular}
    }
    \caption{Inductive bias (in bits) of various combinations of models and datasets under a neural network hypothesis space. Inductive bias of models is assessed using error rates reported in prior literature~\protect\cite{lecun1998mnist,mrgrhn2021alexnet,lee2015deeply,ciregan2012multi,nishimoto2018linear,lin2015far,krizhevsky2012imagenet,huang2017densely,mauch2017prune,veeramacheneni2022canonical,goodfellow2013multi,finn2017model,rajeswaran2019meta,vinyals2016matching,snell2017prototypical,lee2018gradient}. - indicates that results are not available in prior literature.}
    \label{tab:model_comp}
\end{table*}
Second, we consider the hypothesis space expressible by a high-capacity ReLU-activated fully-connected neural network with $9$ layers and $512$ units per hidden layer. The appendix describes the details of how the hypothesis space is constructed. We sample hypotheses from this space by training networks of this architecture under different random initialization and training data permutations.  For both spaces, we fit a Chi-squared distribution to the distribution of test errors to compute inductive bias (Section~\ref{chi-squared}). The appendix includes further experimental details.

Table~\ref{tab:bitsresults} shows our inductive bias estimates and compares them to \cite{boopathy2023model}'s prior upper bounds on the inductive bias; \cite{boopathy2023model} use a similarly high-dimensional hypothesis space as our kernel-based hypothesis space.  Under both our hypothesis spaces, our measure is \textit{many orders of magnitude} lower than \cite{boopathy2023model}'s prior upper bound. This is because \cite{boopathy2023model} computed upper bounds of inductive bias, while we used a more precise estimation method. Further, our hypothesis space, although quite broad, is different than theirs and potentially more restricted, leading to fewer bits being needed to narrow down the well-generalizing hypotheses.

We also see that tasks with more intrinsic dimensionality require more bits of inductive bias to generalize well; in particular, Inverted Pendulum $<$ MNIST $<$ CIFAR-10 $<$ Omniglot, which matches expectations from \cite{boopathy2023model}. Moreover, under the neural network hypothesis space, the required inductive bias is lower than for the kernel hypothesis space, especially for the simpler tasks; for all tasks but Omniglot, the inductive bias is roughly an order of magnitude lower. Thus, neural networks provide a strong bias compared to kernels for these tasks. On the other hand, since Omniglot is a few-shot learning task, plain neural networks provide limited inductive bias for it relative to kernels. Specialized few-shot learning methods include the inductive bias that the output should relate to the query input in the same way that the training labels relate to the training inputs: in other words, they know the \textit{analogical structure} of the task. Plain neural networks do not contain this information, and thus carry limited inductive bias in this setting.

\subsection{Inductive Bias in Different Models}
Next, we evaluate the amount of inductive bias \textit{contained} by different models. Note that our inductive bias measure is a function of the desired test set error rate ($\varepsilon$ in Equation~\ref{eqn:definition}); previously, we set the desired error rate as a fixed value for each task. However, following~\cite{boopathy2023model}, we may also compute the inductive bias \textit{provided} by different models by evaluating the test set error of the models and plugging this error into $\varepsilon$ in Equation~\ref{eqn:definition}. The inductive bias provided by a model is the amount of information required to achieve the error rate of the model. We evaluate the inductive bias contained models trained in prior work. We use the high-capacity neural network hypothesis space described previously. Note that the architecture of the model used to construct the hypothesis space may be smaller than some of the architectures we evaluate; nevertheless, we believe the hypothesis space is expressive enough to contain functions sufficiently close to the trained models in prior literature. See the appendix for additional evaluation details. 

In Table~\ref{tab:model_comp}, we find that the inductive bias provided by different models trained on the same task is similar. Intuitively, this is because for models to generalize similarly, they must provide similar levels of inductive bias. We also observe that the inductive bias of several simple models is $0$ bits. This corresponds to random models in the hypothesis space outperforming the evaluated model; thus, no \textit{additional} information is provided by the inductive biases in the model relative to the hypothesis space.

\section{Discussion}

Our results reveal that different tasks require different levels of inductive bias, with higher dimensional tasks demanding greater amounts. In particular, with expressive kernel-based hypothesis spaces, the required inductive bias can be higher for high-dimensional tasks such a Omniglot compared to lower-dimensional tasks such as CIFAR-10 \textit{even when the lower-dimensional task may be intuitively simpler}. This curse of dimensionality occurs due to an exponential explosion of the size of the hypothesis space with the task dimension: intuitively, each additional dimension of variation in a task increases the \textit{dimensionality} of the hypothesis space by a constant factor. Our findings confirm previous research and highlight the importance of the choice of model class, particularly for high-dimensional problems.

We also find that neural networks as a model class, inherently encode large amounts of inductive bias. The choice of neural networks themselves provides a much greater inductive bias than specific architectural choices, although our measure also reveals that architectural choices can provide significant inductive bias. This observation suggests that the strong smoothness~\cite{li2018visualizing} and compositionality~\cite{mhaskar2017when} constraints of neural networks align well with the properties of realistic tasks. Consequently, these models naturally embody the inductive bias required for a wide range of tasks, underscoring their prevalence and success across various domains.

We note that our empirical results are restricted to two specific choices of hypothesis spaces: a Gaussian RBF kernel-based hypothesis space and a fixed neural network hypothesis space. However, our approach is applicable to \textit{general} hypothesis spaces. For instance, neural network hypothesis spaces may be constructed in alternate ways than  optimization by gradient descent. For instance, we may train a single neural network with dropout, then drop out different sets of nodes in the network with each set corresponding to a sample from the hypothesis space. Future work may be able to extend our inductive bias quantification to these settings.

We propose two potential ways of using our inductive bias quantification. First, it provides an information-theoretic interpretation of the advantages of particular model architectures for specific tasks. By quantifying the amount of inductive bias associated with different architectural choices, researchers can gain insights into how specific design decisions affect the model's ability to generalize. This understanding helps identify which architectural features contribute most significantly to improved performance and informs the development of more tailored and task-specific models.

Second, the inductive bias measure serves as a quantitative guide for developing tasks that require greater inductive bias. By precisely estimating the amount of inductive bias needed for a given task, researchers can intentionally design challenging benchmarks that push the boundaries of machine learning capabilities. We hope this can encourage the development of more powerful model architectures and learning algorithms that drive the field forward.

\bibliographystyle{named}
\bibliography{ijcai24}

\appendix

\onecolumn
\section{Proof of Theorem 1} \label{app:proof}
\begin{proof}
    Denote the probability distribution over $\mathbb{E}_{x \sim p_x}[L(h, x)]$ when $h$ is drawn from $p_h$ as $p_l(l)$. From the definition of $\tilde I$ and $I(\varepsilon, p_h, p_x, L)$, it is known that:
    \begin{equation}
        |\tilde I - I(\varepsilon, p_h, p_x, L)| = |-\log \int_{-\infty}^{\varepsilon} f(l;\alpha^*) dl + \log \int_{-\infty}^{\varepsilon} p_l(l) dl|
    \end{equation}
    Expressing the difference of logs as a log of a ratio:
    \begin{equation}
    |\tilde I - I(\varepsilon, p_h, p_x, L)| = \left|-\log \frac{\int_{-\infty}^{\varepsilon} f(l;\alpha^*) dl}{\int_{-\infty}^{\varepsilon} p_l(l) dl}\right|
    \end{equation}
    Observe that this expression can be upper bounded by:
    \begin{equation}
        \max_l |\log f(l;\alpha^*) - \log p_l(l)|
    \end{equation}
    To see this, denote $k=\max_l |\log f(l;\alpha^*) - \log p_l(l)|$. Then,
    \begin{equation}
        f(l;\alpha^*) \leq e^{k} p_l(l)
    \end{equation}
    This implies:
    \begin{equation}
        \log \frac{\int_{-\infty}^{\varepsilon} f(l;\alpha^*) dl}{\int_{-\infty}^{\varepsilon} p_l(l) dl} \leq \log \frac{\int_{-\infty}^{\varepsilon} e^k p_l(l) dl}{\int_{-\infty}^{\varepsilon} p_l(l) dl} = \log e^k \frac{\int_{-\infty}^{\varepsilon}  p_l(l) dl}{\int_{-\infty}^{\varepsilon} p_l(l) dl} = k
    \end{equation}
    Similarly, $-\log \frac{\int_{-\infty}^{\varepsilon} f(l;\alpha^*) dl}{\int_{-\infty}^{\varepsilon} p_l(l) dl}$ can be upper bounded by $k$. Thus,
    \begin{equation}
        |\tilde I - I(\varepsilon, p_h, p_x, L)| \leq \max_l |\log f(l;\alpha^*) - \log p_l(l)| 
    \end{equation}
    We can upper bound the absolute value as:
    \begin{equation}
        \max_l |\log f(l;\alpha^*) - \log p_l(l)| \leq \max_l |\log f(l;\alpha^*) - \log q_l(l)| + \max_l |\log q_l(l) - \log p_l(l)|
    \end{equation}
    We first bound the first term. Reparameterizing $f$ in terms of moments $\mu$:
    \begin{equation}
        \max_l |\log f(l;\alpha^*) - \log q_l(l)| = \max_l |\log \tilde f(l;\mu^*) - \log q_l(l)|
    \end{equation}
    where $\mu^*$ are moment estimates computed as sample averages. We upper bound the absolute value as:
    \begin{equation}
        \max_l |\log \tilde f(l;\mu^*) - \log q_l(l)| \leq \max_l |\log \tilde f(l;\mu^*) - \log \tilde f(l;\bar \mu)| + \max_l |\log \tilde f(l;\bar \mu) - \log q_l(l)|
    \end{equation}
    Note that $\max_l |\log \tilde f(l;\bar \mu) - \log q_l(l)|$ is bounded by $\xi_l$. Since $f$ has bounded support, by Hoeffding's inequality, the deviation from the mean of a single element of $\mu^*$ can be bounded as:
    \begin{equation}
        P(|\mu_i^* - \bar \mu_i|^2 \geq \frac{t}{r}) \leq 2 e^{- C \frac{n^2 t}{r}}
    \end{equation}
    for some constant $C$. Using the union bound:
    \begin{equation}
        P(||\mu_i^* - \bar \mu_i||^2 \geq t) \leq 2 r e^{- C \frac{n^2 t}{r}}
    \end{equation}
    We set $\sigma = 2 r e^{-C \frac{n^2 t}{r}}$, which yields:
    \begin{equation}
        t = \frac{r}{Cn^2} \log \frac{2r}{\sigma}
    \end{equation}
    Thus, with probability $1 - \sigma$:
    \begin{equation}
        P(||\mu_i^* - \bar \mu_i|| \leq \frac{1}{n} \sqrt{\frac{r}{C} \log \frac{2r}{\sigma}})
    \end{equation}
    Since $\log \tilde f$ is Lipschitz continuous with respect to $\mu$, with probability $1 - \sigma$:
    \begin{equation}
        \max_l |\log \tilde f(l;\mu^*) - \log \tilde f(l;\bar \mu)| \leq \frac{\kappa}{n} \sqrt{r \log \frac{2r}{\sigma}}
    \end{equation}
    for some constant $\kappa$. Thus, we may bound $\max_l |\log f(l;\alpha^*) - \log q_l(l)|$ as:
    \begin{equation}
        \max_l |\log f(l;\alpha^*) - \log q_l(l)| \leq \xi_l + \frac{\kappa}{n} \sqrt{r \log \frac{2r}{\sigma}}
    \end{equation}
    Next, we bound $\max_l |\log q_l(l) - \log p_l(l)|$ using the bound on $\max_h |\log q_h(h) - \log p_h(h)|$. Note that
    \begin{equation}
        \log q_l(l) - \log p_l(l) = \log \frac{\int_{h: \mathbb{E}_{x \sim p_x}[L(h,x)] = l} q_h(h) dh}{\int_{h: \mathbb{E}_{x \sim p_x}[L(h,x)] = l} p_h(h) dh} \leq \log \frac{\int_{h: \mathbb{E}_{x \sim p_x}[L(h,x)] = l} e^{\xi_h} p_h(h) dh}{\int_{h: \mathbb{E}_{x \sim p_x}[L(h,x)] = l} p_h(h) dh} = \xi_h
    \end{equation}
    Similarly,
    \begin{equation}
        \log p_l(l) - \log q_l(l) = \log \frac{\int_{h: \mathbb{E}_{x \sim p_x}[L(h,x)] = l} p_h(h) dh}{\int_{h: \mathbb{E}_{x \sim p_x}[L(h,x)] = l} q_h(h) dh} \leq \log \frac{\int_{h: \mathbb{E}_{x \sim p_x}[L(h,x)] = l} e^{\xi_h} q_h(h) dh}{\int_{h: \mathbb{E}_{x \sim p_x}[L(h,x)] = l} q_h(h) dh} = \xi_h
    \end{equation}
    Therefore,
    \begin{equation}
        \max_l |\log q_l(l) - \log p_l(l)| \leq \xi_h
    \end{equation}
    Combining all the inequalities, with probability $1-\sigma$:
    \begin{equation}
        |\tilde I - I(\varepsilon, p_h, p_x, L)| \leq \xi_h + \xi_l + \frac{\kappa}{n} \sqrt{r \log \frac{2r}{\sigma}}
    \end{equation}
\end{proof}

\section{Additional Experimental Details} \label{app:exps}

\subsection{Kernel-based Sampling Details} \label{app:exps_kernel}

Using the decomposition of $\theta$, $h(\bar X) \in \mathbb{R}^{nk}$ can be expressed as:
\begin{equation}
    h(\bar X) = \phi(\bar X) \phi(X)^{\dagger} Y + \phi(\bar X) \beta
\end{equation}
where $\phi(\bar X) \in \mathbb{R}^{nk \times P}$. Note that this is distributed as a Gaussian with mean $\phi(\bar X) \phi(X)^{\dagger} Y$ and covariance matrix $\phi(\bar X) (I - \phi(X)^\dagger \phi(X)) \phi(\bar X)^T = \phi(\bar X) \phi(\bar X)^T - \phi(\bar X) \phi(X)^\dagger \phi(X) \phi(\bar X)^T$. We may express these quantities in terms of the \textit{kernel} corresponding to features $\phi(x)$. The kernel is defined a $k \times k$ matrix:
\begin{equation}
    K(x_1, x_2) = \phi(x_1) \phi(x_2)^T
\end{equation}
In our experiments, we use a Gaussian radial basis function (RBF) kernel. We denote $K(X, X) \in \mathbb{R}^{Nk, Nk}$ and $K(\bar X, X) \in \mathbb{R}^{nK, Nk}$ as the kernels between all pairs of training points and pairs of test and training points respectively. Then, assuming $N < P$, $h(\bar X)$ has mean $K(\bar X, X) K(X, X)^{-1} Y$ and covariance $K(\bar X, \bar X) - K(\bar X, X) K(X, X)^{-1} K(X, \bar X)$. Thus, we may express $h(\bar X)$ as:
\begin{equation}
    h(\bar X) = K(\bar X, X) K(X, X)^{-1} Y + \sqrt{K(\bar X, \bar X) - K(\bar X, X) K(X, X)^{-1} K(X, \bar X)} z
\end{equation}
where $z \in \mathbb{R}^{nk}$ is distributed as a unit Gaussian and $\sqrt{}$ denotes matrix square root. Computing this quantity directly for a given choice of $z$ may be computationally challenging since it requires storing and inverting the kernel matrix $K(X, X)$. Thus, we instead approximate solutions to $K(X, X)^{-1} Y$ and $K(X, X)^{-1} K(X, \bar X)$ via gradient descent. Specifically, we define $\alpha^* = K(X, X)^{-1} Y$ and $A^* = K(X, X)^{-1} K(X, \bar X)$. Note that these can be found as solutions to the following optimization problems:
\begin{equation}
    \alpha^* = \argmin_\alpha ||Y - K(X, X) \alpha||_2^2
\end{equation}
\begin{equation}
    A^* = \argmin_A ||K(X, \bar X) - K(X, X) A||_F^2
\end{equation}
We find approximate solutions to these optimization problems by stochastic gradient descent. Once $\alpha^*$ and $A^*$ are found, samples of $h(\bar X)$ may be computed as:
\begin{equation}
    h(\bar X) = K(\bar X, X) \alpha^* + \sqrt{K(\bar X, \bar X) - K(\bar X, X) A^*} z
\end{equation}
Assuming a constant time kernel computation per element, performing stochastic gradient descent on $\alpha$ and $A$ for one iteration requires $O(Nk)$ and $O(nNk^2)$ time respectively; thus, for $T$ optimization steps, the time complexity is $O(nNk^2T)$. Once computed, sampling $h(\bar X)$ requires $O(n^3k^3 + n^2Nk^3)$ time to determine the constants in the above equation, and an additional $O(n^2k^2)$ per sample. Thus, producing $S$ samples requires a total of $O(nNk^2T + n^3k^3 + n^2Nk^3 + n^2k^2S)$ time. Importantly, this time is \textit{linear} in the training set size and \textit{constant} in the parameter size; this efficiency is critical for large datasets and high-dimensional hypothesis spaces.

\begin{wrapfigure}{r}{0.45\textwidth}
    \centering
    \includegraphics[width=0.35\textwidth]{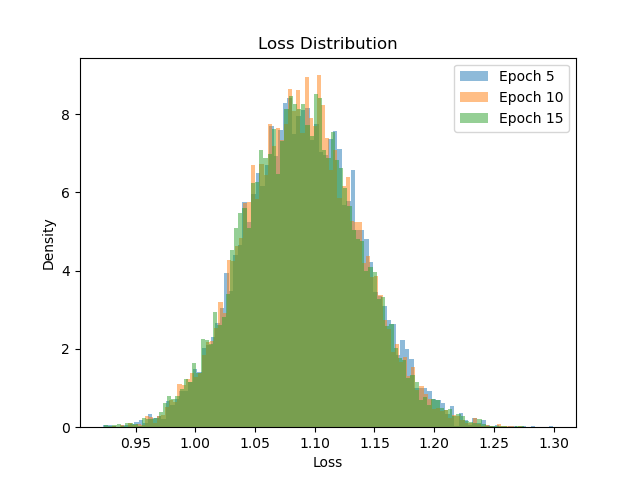}
    \caption{Distribution of hypothesis losses for MNIST after 5, 10, and 15 epochs of gradient descent. Notice that the update in distribution is minimal.}
    \label{fig:mnistconvergence}
\end{wrapfigure}

We use a Gaussian RBF kernel constructed as $K(x_1, x_2) = e^{-\frac{1}{2} ||x_1 - x_2||_2^2 I}$. Due to the computational cost of computing the full gradients $g_\alpha$ and $g_{A}$ (both of which have sizes scaling linearly with the training set size), we compute the gradient in steps. Specifically, for $g_{\alpha}$, $K(x, X) \alpha$ is computed by splitting the training set into groups $X_i$ (and correspondingly $\alpha$ into $\alpha_i$) and summing the contribution of each $K(x, X_i) \alpha_i$. An analogous grouping is done to compute $g_A$.

For experiments on MNIST, we use a learning rate of $0.0001$ to optimize $\alpha$ and a learning rate of $0.00001$ to optimize $A$. We use a batch size of $128$ and a group size of $2048$. Training is performed for $20$ epochs. See Figure~\ref{fig:mnistconvergence} for an analysis of the convergence of the resulting loss distribution.

For experiments on CIFAR-10, we use a learning rate of $0.001$ to optimize $\alpha$ and a learning rate of $0.0001$ to optimize $A$. We use a batch size of $64$ and a group size of $1024$. Training is performed for $20$ epochs.

% For experiments on ImageNet,

For experiments on Omniglot, we use a learning rate of $0.0001$ to optimize $\alpha$ and a learning rate of $0.00001$ to optimize $A$. We use a batch size of $10$ and a group size of $100$. Training is performed for $1$ epoch.

For experiments on the inverted pendulum task, we use a learning rate of $0.001$ to optimize $\alpha$ and a learning rate of $0.0001$ to optimize $A$. We use a batch size of $64$ and a group size of $1024$. Training is performed for $500$ epochs.

For the kernel-based hypothesis space, we sample a total of $100000$ hypotheses in each setting.

\subsection{Neural Network Hypothesis Space Details}
The architecture of our base hypothesis space is constructed as follows: the input is linearly projected to a $512$ dimensional vector, followed by $9$ more fully connected layers of dimensionality $512$. Finally, the output is linearly projected to a $10$-dimensional output to predict the one-hot encoded label. All layers include a bias term. Each fully connected layer except the final one is followed by a ReLU non-linearity. No additional components such as normalization are used.

The model is trained on a mean squared error loss using Adam with a learning rate of $0.001$. Training is conducted for $10$ epochs with a batch size of $128$. We sample $100$ hypotheses from this hypothesis space by using different random initializations and orderings of training points during training.

Within the neural network hypothesis space, we consider the performance of models trained in prior work. Prior work typically reports the test set accuracy of each model; however, our method uses test set mean squared error as an error metric instead since the theoretical prediction of a Chi-squared error distribution is only available for mean-squared error. Thus, we estimate mean squared errors from accuracies by simply setting mean squared error $e$ as $1-a$ where $a$ represents accuracy.

\begin{wrapfigure}{r}{0.4\textwidth}
    \centering
    \includegraphics[width=0.4\textwidth]{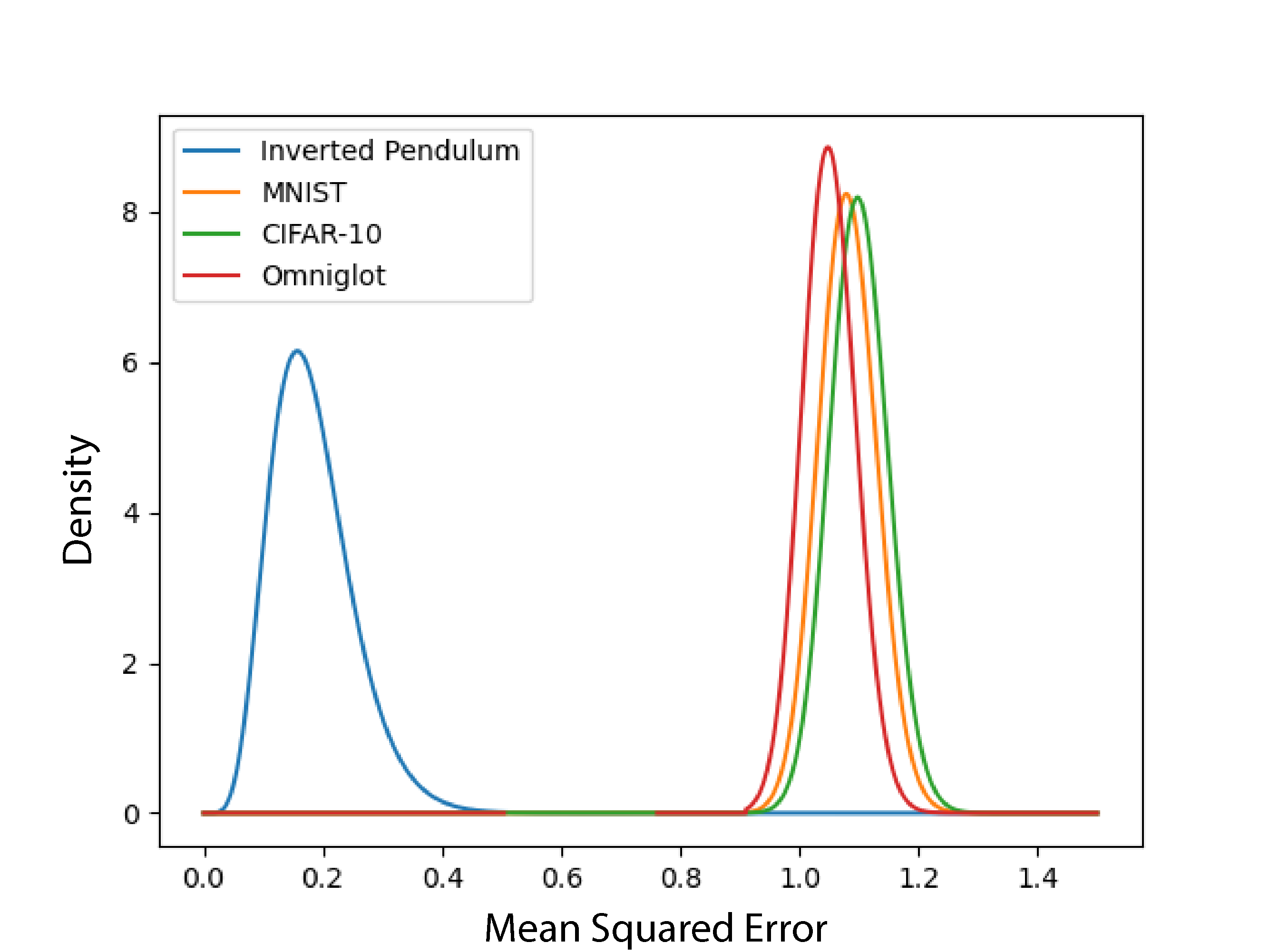}
    \caption{Fitted scaled non-central Chi-squared distributions for the test set errors on MNIST, CIFAR-10, Omniglot, and Inverted Pendulum tasks under a Gaussian RBF kernel hypothesis space.}
    \label{fig:fittedchisquaredresults}
    \vspace{-8mm}
\end{wrapfigure}
\subsection{Distribution Fitting Details} \label{app:exps_dist}
Given samples from the hypothesis space, compute the test set error for each one. Then, given the empirical distribution of test set errors, we fit a three-parameter scaled non-central Chi-squared distribution to match the data. We use maximum likelihood estimation to determine the fit parameters. Figure~\ref{fig:fittedchisquaredresults} illustrates example fitted distributions under the Gaussian RBF kernel hypothesis space. Once the distribution parameters are determined, we use approximations to quantify the log of the cumulative distribution function as described in the main text.

\subsection{Tasks}
For all tasks, desired error rates are set as the values provided in~\cite{boopathy2023model} unless otherwise specified.

\paragraph{MNIST \& CIFAR-10}
We use the base MNIST and CIFAR-10 datasets without modification.

\paragraph{ImageNet}
We use standard ImageNet normalization and random cropping.

\paragraph{Omniglot}
We consider 5-way 1-shot Omniglot classification. In this setting, each input consists of $5$ images, $1$ from each of $5$ alphabets, and the goal is to predict the class of a new image from one of the $5$ seen alphabets. We encode this task in the following form: the inputs $x$ consist of $6$ images, the first $5$ of which correspond to training images and the last one of which corresponds to the evaluation image. The input is flattened to remove all spatial structure. The desired output is a one-hot encoded $5$-dimensional vector of which of the $5$ training images matches the class of the evaluation image. We generate a training set of size $10000$ and a test set of size $100$, drawn from the Omniglot background and evaluation alphabets respectively.

\paragraph{Inverted Pendulum Task}
We consider the following inverted pendulum control task: an inverted pendulum with angle $\theta \in \mathbb{R}$ and angular velocity $\omega \in \mathbb{R}$ has the following dynamics:
\begin{equation}
    \dot \theta = \omega
\end{equation}
\begin{equation}
    \dot \omega = \sin \theta + u
\end{equation}
where $u \in \mathbb{R}$ is a control action. The goal is to minimize the time average of the following cost:
\begin{equation}
    C(u, \theta, \omega) = \frac{1}{2} u^2 + 24 \theta^2  + (8 \theta + 4\omega) (\theta - \sin \theta)
\end{equation}
The first term encourages small control actions. The second term encourages the inverted pendulum to remain at rest at $\theta=0$. The third term is added to allow an analytically tractable optimal control. Intuitively, it penalizes when future values of $\theta$ (as represented by $8\theta + 4\omega$) are far from $0$ (as represented by $\theta - \sin \theta$). Note that this term is on the order of $O(\theta^3)$; for small $\theta$, the second term dominates.

The optimal cost to go, or value function (i.e. the optimal total cost over all future time steps) given current state $\theta, \omega$ is:
\begin{equation}
    V(\theta, \omega) = 14 \theta^2 + 8 \theta \omega + 2 \omega^2
\end{equation}
To verify this, note that the optimal cost obeys the Bellman equation:
\begin{equation}
    \min_u \frac{\partial}{\partial \theta} V(\theta, \omega) \dot \theta + \frac{\partial}{\partial \omega} V(\theta, \omega) \dot \omega + C(u, \theta, \omega) = 0
\end{equation}
Substituting in the expressions from above:
\begin{equation}
    \min_u [(28 \theta + 8 \omega) \omega + (8 \theta + 4 \omega)(\sin \theta + u) + \frac{1}{2} u^2 + 24 \theta^2  + (8 \theta + 4\omega) (\theta - \sin \theta)]= 0
\end{equation}
Setting the derivative with respect to $u$ to $0$, the optimal $u$ must satisfy:
\begin{equation}
    8 \theta + 4 \omega + u = 0
\end{equation}
Thus, $u = -8 \theta - 4 \omega$. Plugging this back into the Bellman equation:
\begin{equation}
    (28 \theta + 8 \omega) \omega + (8 \theta + 4 \omega)(\sin \theta -8 \theta - 4 \omega) + \frac{1}{2} (-8 \theta - 4 \omega)^2 + 24 \theta^2  + (8 \theta + 4 \omega) (\theta - \sin \theta) = 0
\end{equation}
Observe that all terms on the left-hand side cancel; thus $V(\theta, w)$i is the correct value function for this optimal control problem. Given $\theta, \omega$, the optimal control action is:
\begin{equation}
    u = -8 \theta - 4 \omega
\end{equation}

In this task, inputs $(\theta, \omega)$ are drawn from a uniform distribution over $[-\pi, \pi] \times [-1, 1]$. Desired outputs $u$ are constructed as above. We generate a training set of size $10000$ and a test set of size $100$.

\subsection{Computing Infrastructure}
Experiments are run on a computing cluster with GPUs ranging in memory size from 11 GB to 80 GB.

\end{document}